\documentclass{article}

\usepackage{microtype}
\usepackage{graphicx}
\usepackage{subfigure}
\usepackage{booktabs} % for professional tables
\usepackage[utf8]{inputenc}
\usepackage{multicol}

\usepackage{hyperref}

\usepackage{amsmath,amsfonts,bm}

\def\eqref#1{equation~\ref{#1}}
\def\1{\bm{1}}

\def\vc{{\bm{c}}}

\def\vx{{\bm{x}}}

\def\vz{{\bm{z}}}

\def\mW{{\bm{W}}}

\DeclareMathAlphabet{\mathsfit}{\encodingdefault}{\sfdefault}{m}{sl}
\SetMathAlphabet{\mathsfit}{bold}{\encodingdefault}{\sfdefault}{bx}{n}

\def\sD{{\mathbb{D}}}
\usepackage{listings}
\usepackage{xcolor}

\definecolor{maroon}{cmyk}{0, 0.87, 0.68, 0.32}
\definecolor{halfgray}{gray}{0.55}
\definecolor{ipython_frame}{RGB}{207, 207, 207}
\definecolor{ipython_bg}{RGB}{247, 247, 247}
\definecolor{ipython_red}{RGB}{186, 33, 33}
\definecolor{ipython_green}{RGB}{0, 128, 0}
\definecolor{ipython_cyan}{RGB}{64, 128, 128}
\definecolor{ipython_purple}{RGB}{170, 34, 255}

\lstdefinelanguage{Python}{
    morekeywords={access,and,break,class,continue,def,del,elif,else,except,exec,finally,for,from,global,if,import,in,is,lambda,not,or,pass,print,raise,return,try,while},
    morekeywords=[2]{abs,all,any,basestring,bin,bool,bytearray,callable,chr,classmethod,cmp,compile,complex,delattr,dict,dir,divmod,enumerate,eval,execfile,file,filter,float,format,frozenset,getattr,globals,hasattr,hash,help,hex,id,input,int,isinstance,issubclass,iter,len,list,locals,long,map,max,memoryview,min,next,object,oct,open,ord,pow,property,range,raw_input,reduce,reload,repr,reversed,round,set,setattr,slice,sorted,staticmethod,str,sum,super,tuple,type,unichr,unicode,vars,xrange,zip,apply,buffer,coerce,intern},
    sensitive=true,
    morecomment=[l]\#,
    morestring=[b]',
    morestring=[b]",
    morestring=[s]{'''}{'''},
    morestring=[s]{"""}{"""},
    morestring=[s]{r'}{'},
    morestring=[s]{r"}{"},
    morestring=[s]{r'''}{'''},
    morestring=[s]{r"""}{"""},
    morestring=[s]{u'}{'},
    morestring=[s]{u"}{"},
    morestring=[s]{u'''}{'''},
    morestring=[s]{u"""}{"""},
    literate=
    {á}{{\'a}}1 {é}{{\'e}}1 {í}{{\'i}}1 {ó}{{\'o}}1 {ú}{{\'u}}1
    {Á}{{\'A}}1 {É}{{\'E}}1 {Í}{{\'I}}1 {Ó}{{\'O}}1 {Ú}{{\'U}}1
    {à}{{\`a}}1 {è}{{\`e}}1 {ì}{{\`i}}1 {ò}{{\`o}}1 {ù}{{\`u}}1
    {À}{{\`A}}1 {È}{{\'E}}1 {Ì}{{\`I}}1 {Ò}{{\`O}}1 {Ù}{{\`U}}1
    {ä}{{\"a}}1 {ë}{{\"e}}1 {ï}{{\"i}}1 {ö}{{\"o}}1 {ü}{{\"u}}1
    {Ä}{{\"A}}1 {Ë}{{\"E}}1 {Ï}{{\"I}}1 {Ö}{{\"O}}1 {Ü}{{\"U}}1
    {â}{{\^a}}1 {ê}{{\^e}}1 {î}{{\^i}}1 {ô}{{\^o}}1 {û}{{\^u}}1
    {Â}{{\^A}}1 {Ê}{{\^E}}1 {Î}{{\^I}}1 {Ô}{{\^O}}1 {Û}{{\^U}}1
    {œ}{{\oe}}1 {Œ}{{\OE}}1 {æ}{{\ae}}1 {Æ}{{\AE}}1 {ß}{{\ss}}1
    {ç}{{\c c}}1 {Ç}{{\c C}}1 {ø}{{\o}}1 {å}{{\r a}}1 {Å}{{\r A}}1
    {€}{{\EUR}}1 {£}{{\pounds}}1
    {^}{{{\color{ipython_purple}\^{}}}}1
    {=}{{{\color{ipython_purple}=}}}1
    {+}{{{\color{ipython_purple}+}}}1
    {*}{{{\color{ipython_purple}$^\ast$}}}1
    {/}{{{\color{ipython_purple}/}}}1
    {+=}{{{+=}}}1
    {-=}{{{-=}}}1
    {*=}{{{$^\ast$=}}}1
    {/=}{{{/=}}}1,
    literate=
    *{-}{{{\color{ipython_purple}-}}}1
     {?}{{{\color{ipython_purple}?}}}1,
    identifierstyle=\color{black}\ttfamily,
    commentstyle=\color{ipython_cyan}\ttfamily,
    stringstyle=\color{ipython_red}\ttfamily,
    keepspaces=true,
    showspaces=false,
    showstringspaces=false,
    rulecolor=\color{ipython_frame},
    frame=single,
    frameround={t}{t}{t}{t},
    framexleftmargin=6mm,
    numbers=left,
    numberstyle=\tiny\color{halfgray},
    backgroundcolor=\color{ipython_bg},
    basicstyle=\scriptsize,
    keywordstyle=\color{ipython_green}\ttfamily,
}

\usepackage{amsmath}
\usepackage{xcolor}

\newcommand*{\colorboxed}{}
\def\colorboxed#1#{%
  \colorboxedAux{#1}%
}
\newcommand*{\colorboxedAux}[3]{%
  \begingroup
    \colorlet{cb@saved}{.}%
    \color#1{#2}%
    \boxed{%
      \color{cb@saved}%
      #3%
    }%
  \endgroup
}

\usepackage[accepted]{icml2020}

\icmltitlerunning{Data-Efficient Image Recognition  with Contrastive Predictive Coding}

\begin{document}

\twocolumn[
\icmltitle{Data-Efficient Image Recognition with Contrastive Predictive Coding}




\begin{icmlauthorlist}
\icmlauthor{Olivier J. Hénaff}{goo}
\icmlauthor{Aravind Srinivas}{bair}
\icmlauthor{Jeffrey De Fauw}{goo}
\icmlauthor{Ali Razavi}{goo} \\
\icmlauthor{Carl Doersch}{goo}
\icmlauthor{S. M. Ali Eslami}{goo}
\icmlauthor{Aaron van den Oord}{goo}
\end{icmlauthorlist}

\icmlaffiliation{goo}{DeepMind, London, UK}
\icmlaffiliation{bair}{University of California, Berkeley}

\icmlcorrespondingauthor{Olivier J. Hénaff}{henaff@google.com}

\icmlkeywords{Machine Learning, ICML}

\vskip 0.3in
]



\printAffiliationsAndNotice{}  % leave blank if no need to mention equal contribution

\begin{abstract}
Human observers can learn to recognize new categories of images from a handful of examples, yet doing so with artificial ones remains an open challenge. We hypothesize that data-efficient recognition is enabled by representations which make the variability in natural signals more predictable. We therefore revisit and improve Contrastive Predictive Coding, an unsupervised objective for learning such representations. This new implementation produces features which support state-of-the-art linear classification accuracy on the ImageNet dataset. When used as input for non-linear classification with deep neural networks, this representation allows us to use 2--5$\times$ less labels than classifiers trained directly on image pixels. Finally, this unsupervised representation substantially improves transfer learning to object detection on the PASCAL VOC dataset, surpassing fully supervised pre-trained ImageNet classifiers. 
\end{abstract}

\section{Introduction}

\begin{figure}
\centering
    \includegraphics[width=0.47\textwidth]{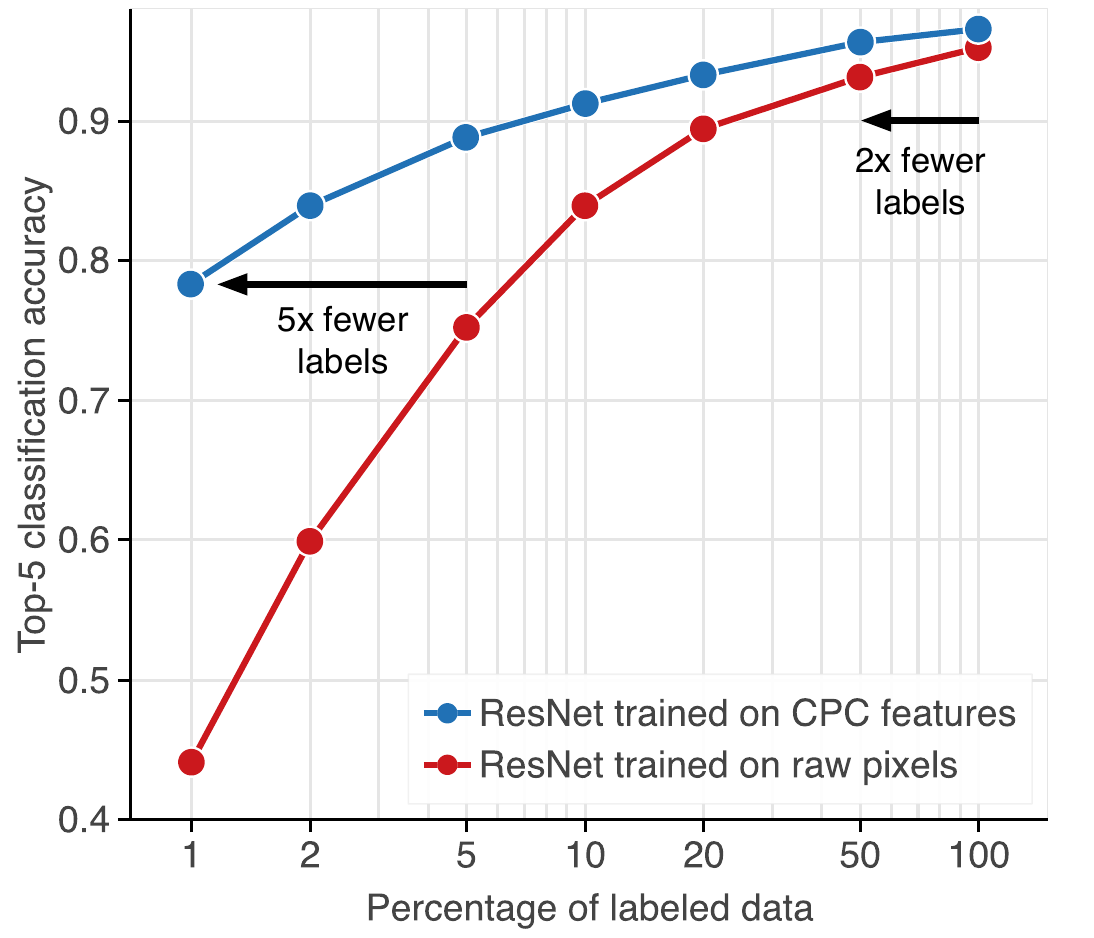}
    \vspace{-1em}
   \caption{Data-efficient image recognition with Contrastive Predictive Coding. With decreasing amounts of labeled data, supervised networks trained on pixels fail to generalize (red). When trained on unsupervised representations learned with CPC, these networks retain a much higher accuracy in this low-data regime (blue). Equivalently, the accuracy of supervised networks can be matched with significantly fewer labels (horizontal arrows).\vspace{-2em}}
\label{fig:intro}
\end{figure}

Deep neural networks excel at perceptual tasks when labeled data are abundant, yet their performance degrades substantially when provided with limited supervision (Fig.\ \ref{fig:intro}, red). In contrast, humans and animals can learn about new classes of images from a small number of examples \citep{landau1988importance, markman1989categorization}. What accounts for this monumental difference in data-efficiency between biological and machine vision? While highly structured representations (e.g.\ as proposed by~\citet{lake2015human}) may improve data-efficiency, it remains unclear how to program explicit structures that capture the enormous complexity of real-world visual scenes, such as those present in the ImageNet dataset~\citep{russakovsky2015imagenet}. An alternative hypothesis has therefore proposed that intelligent systems need not be structured \textit{a priori}, but can instead learn about the structure of the world in an unsupervised manner \citep{barlow1988unsup, hinton1999unsup, lecun2015deep}. Choosing an appropriate training objective is an open problem, 
but a potential guiding principle is that useful representations should make the variability in natural signals more predictable \citep{Tishby1999, wiskott2002slow, Richthofer2016}. Indeed, human perceptual representations have been shown to linearize (or `straighten') the temporal transformations found in natural videos, a property lacking from current supervised image recognition models \citep{henaff2019perceptual}, and theories of both spatial and temporal predictability have succeeded in describing properties of early visual areas \citep{rao1999predictive, palmer2015predictive}. In this work, we hypothesize that spatially predictable representations may allow artificial systems to benefit from human-like data-efficiency.

Contrastive Predictive Coding (CPC, \citet{oord2018representation}) is an unsupervised objective which learns predictable representations. CPC is a general technique that only requires in its definition that observations be ordered along e.g.\ temporal or spatial dimensions, and as such has been applied to a variety of different modalities including speech, natural language and images. This generality, combined with the strong performance of its representations in downstream linear classification tasks, makes CPC a promising candidate for investigating the efficacy of predictable representations for data-efficient image recognition.

Our work makes the following contributions:
\begin{itemize}
    \item We revisit CPC in terms of its architecture and training methodology, and arrive at a new implementation with a dramatically-improved ability to linearly separate image classes (from 48.7\% to 71.5\% Top-1 ImageNet classification accuracy, a 23\% absolute improvement), setting a new state-of-the-art. 
    \item We then train deep neural networks on top of the resulting CPC representations using very few labeled images (e.g.\ 1\% of the ImageNet dataset), and demonstrate test-time classification accuracy far above networks trained on raw pixels (78\% Top-5 accuracy, a 34\% absolute improvement), outperforming all other semi-supervised learning methods (+20\% Top-5 accuracy over the previous state-of-the-art \citep{zhai2019s}). This gain in accuracy allows our classifier to surpass supervised ones trained with 5$\times$ more labels. 
    \item Surprisingly, this representation also surpasses supervised ResNets when given the entire ImageNet dataset (+3.2\% Top-1 accuracy). Alternatively, our classifier is able to match fully-supervised ones while only using half of the labels.
    \item Finally, we assess the generality of CPC representations by transferring them to a new task and dataset: object detection on PASCAL VOC 2007. Consistent with the results from the previous sections, we find CPC to give state-of-the-art performance in this setting (76.6\% mAP), surpassing the performance of supervised pre-training (+2\% absolute improvement). 
\end{itemize}

\section{Experimental Setup}

We first review the CPC architecture and learning objective in section \ref{sec:cpc}, before detailing how we use its resulting representations for image recognition tasks in section \ref{sec:evaluation-protocol}. 

\subsection{Contrastive Predictive Coding}
\label{sec:cpc}

\begin{figure*}[t]
    \includegraphics[width=\textwidth]{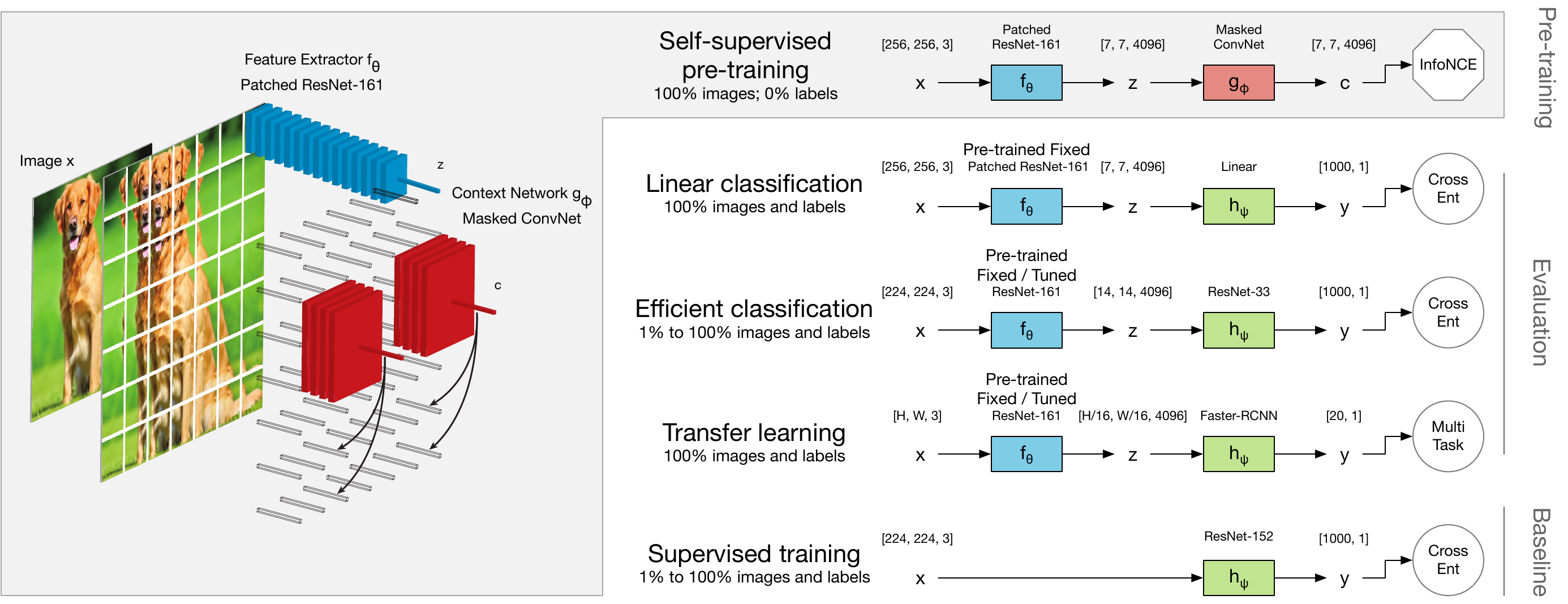}
    \vspace{-2em}
   \caption{
   Overview of the framework for semi-supervised learning with Contrastive Predictive Coding. Left: unsupervised pre-training with the spatial prediction task (See Section \ref{sec:cpc}). First, an image is divided into a grid of overlapping patches. Each patch is encoded independently from the rest with a feature extractor (blue) which terminates with a mean-pooling operation, yielding a single feature vector for that patch. Doing so for all patches yields a field of such feature vectors (wireframe vectors). Feature vectors above a certain level (in this case, the center of the image) are then aggregated with a context network (red), yielding a row of context vectors which are used to linearly predict features vectors below. Right: using the CPC representation for a classification task. Having trained the encoder network, the context network (red) is discarded and replaced by a classifier network (green) which can be trained in a supervised manner. In some experiments, we also fine-tune the encoder network (blue) for the classification task. When applying the encoder to cropped patches (as opposed to the full image) we refer to it as a \textit{patched} ResNet in the figure.} 
\label{fig:cpc}
\end{figure*}

Contrastive Predictive Coding as formulated in \citep{oord2018representation} learns representations by training neural networks to predict the representations of future observations from those of past ones. When applied to images, CPC operates by predicting the representations of patches below a certain position from those above it (Fig.\ \ref{fig:cpc}, left). These predictions are evaluated using a contrastive loss \citep{chopra2005learning, hadsell2006dimensionality}, in which the network must correctly classify `future' representations among a set of unrelated `negative' representations. This avoids trivial solutions such as representing all patches with a constant vector, as would be the case with a mean squared error loss.

In the CPC architecture, each input image is first divided into a grid of overlapping patches $\vx_{i,j}$, where $i,j$ denote the location of the patch. Each patch is encoded with a neural network $f_\theta$ into a single vector $\vz_{i,j}=f_\theta(\vx_{i,j})$. To make predictions, a masked convolutional network $g_\phi$ is then applied to the grid of feature vectors. The masks are such that the receptive field of each resulting \textit{context vector} $c_{i,j}$ only includes feature vectors that lie above it in the image (i.e.\ $ c_{i,j} = g_\phi(\{\vz_{u,v}\}_{ u \leq i, v })$). The prediction task then consists of predicting `future' feature vectors $z_{i+k,j}$ from current context vectors $c_{i,j}$, where $k>0$. The predictions are made linearly: given a context vector $c_{i,j}$, a prediction length $k>0$, and a prediction matrix $W_k$, the predicted feature vector is $ \hat{\vz}_{i+k,j} = \mW_{k}\vc_{i,j} $. 

The quality of this prediction is then evaluated using a contrastive loss. Specifically, the goal is to correctly recognize the target $z_{i+k,j}$ among a set of randomly sampled feature vectors $\{z_l\}$ from the dataset. We compute the probability assigned to the target using a softmax, and rate this probability using the usual cross-entropy loss. Summing this loss over locations and prediction offsets, we arrive at the CPC objective as defined in \citep{oord2018representation}: 
\begin{align*}
\mathcal{L}_\textrm{CPC} 
& = -\sum_{i,j,k} \log p(\vz_{i+k,j} | \hat{\vz}_{i+k,j}, \{\vz_l\} ) \\ 
& = - \sum_{i,j,k} \log \frac{\exp(\hat{\vz}_{i+k,j}^T \vz_{i+k,j})}{\exp(\hat{\vz}_{i+k,j}^T \vz_{i+k,j})+\sum_l \exp(\hat{\vz}_{i+k,j}^T \vz_{l})}
\end{align*}
The \emph{negative samples} $\{\vz_{l}\}$ are taken from other locations in the image and other images in the mini-batch. This loss is called InfoNCE as it is inspired by Noise-Contrastive Estimation \citep{gutmann2010noise, mnih2013learning} and has been shown to maximize the mutual information between $\vc_{i,j}$ and $\vz_{i+k,j}$ \citep{oord2018representation}. 

\subsection{Evaluation protocol}
\label{sec:evaluation-protocol}

Having trained an encoder network $f_\theta$, a context network $g_\phi$, and a set of linear predictors $\{\mW_{k}\}$ using the CPC objective, we use the encoder to form a representation $\vz=f_\theta(\vx)$ of new observations $\vx$, and discard the rest. Note that while pre-training required that the encoder be applied to patches, for downstream recognition tasks we can apply it directly to the entire image. We train a model $h_\psi$ to classify these representations: given a dataset of $N$ unlabeled images $\sD_u = \{x_n\}$, and a (potentially much smaller) dataset of $M$ labeled images $\sD_l = \{x_m, y_m\}$
\begin{align*}
\theta^* & = \arg \min_\theta \frac{1}{N} \sum_{n=1}^N \mathcal{L}_{\textrm{CPC}} [ f_\theta (x_n) ] \qquad \\
\psi^* & = \arg \min_\psi \frac{1}{M} \sum_{m=1}^M \mathcal{L}_{\textrm{Sup}} [ h_\psi \circ f_{\theta^*} ( x_m ) , y_m ]     
\end{align*}
In all cases, the dataset of unlabeled images $\sD_u$ we pre-train on is the full ImageNet ILSVRC 2012 training set \citep{russakovsky2015imagenet}. We consider three labeled datasets $\sD_l$ for evaluation, each with an associated classifier $h_\psi$ and supervised loss $\mathcal{L}_{\textrm{Sup}}$ (see Fig.\ \ref{fig:cpc}, right). This protocol is sufficiently generic to allow us to later compare the CPC representation to other methods which have their own means of learning a feature extractor $f_\theta$. 

\textbf{Linear classification} is a standard benchmark for evaluating the quality of unsupervised image representations. In this regime, the classification network $h_\psi$ is restricted to mean pooling followed by a single linear layer, and the parameters of $f_\theta$ are kept fixed. The labeled dataset $\sD_l$ is the entire ImageNet dataset, and the supervised loss $\mathcal{L}_{\textrm{Sup}}$ is standard cross-entropy. We use the same data-augmentation as in the unsupervised learning phase for training, and none at test time and evaluate with a single crop.

\textbf{Efficient classification} directly tests whether the CPC representation enables generalization from few labels. For this task, the classifier $h_\psi$ is an arbitrary deep neural network (we use an 11-block ResNet architecture~\citep{he2016deep} with 4096-dimensional feature maps and 1024-dimensional bottleneck layers). The labeled dataset $\sD_l$ is a random subset of the ImageNet dataset: we investigated using 1\%, 2\%, 5\%, 10\%, 20\%, 50\% and 100\% of the dataset. The supervised loss $\mathcal{L}_{\textrm{Sup}}$ is again cross-entropy. We use the same data-augmentation as during unsupervised pre-training, none at test-time and evaluate with a single crop.

\textbf{Transfer learning} tests the generality of the representation by applying it to a new task and dataset. For this we chose object detection on the PASCAL VOC 2007 dataset, a standard benchmark in computer vision \citep{everingham2007pascal}. As such $\sD_l$ is the entire PASCAL VOC 2007 dataset (comprised of 5011 labeled images); $h_\psi$ and $\mathcal{L}_{\textrm{Sup}}$ are the Faster-RCNN architecture and loss \citep{ren2015faster}. In addition to color-dropping, we use the  scale-augmentation from \citet{doersch2015unsupervised} for training.

For \textbf{linear classification}, we keep the feature extractor $f_\theta$ \textit{fixed} to assess the representation in absolute terms. For \textbf{efficient classification} and \textbf{transfer learning}, we additionally explore \textit{fine-tuning} the feature extractor for the supervised objective. In this regime, we initialize the feature extractor and classifier with the solutions $\theta^*, \psi^*$ found in the previous learning phase, and train them both for the supervised objective. To ensure that the feature extractor does not deviate too much from the solution dictated by the CPC objective, we use a smaller learning rate and early-stopping.

\section{Related Work}

Data-efficient learning has typically been approached by two complementary methods, both of which seek to make use of more plentiful unlabeled data: representation learning and label propagation. The former formulates an objective to learn a feature extractor $f_\theta$ in an unsupervised manner, whereas the latter directly constrains the classifier $h_\psi$ using the unlabeled data. 

\textbf{Representation learning} saw early success using generative modeling \citep{kingma2014semi}, but likelihood-based models have yet to generalize to more complex stimuli. Generative adversarial models have also been harnessed for representation learning \citep{donahue2016adversarial}, and large-scale implementations have led to corresponding gains in linear classification accuracy \citep{donahue2019large}.

In contrast to generative models which require the reconstruction of observations, self-supervised techniques directly formulate tasks involving the learned representation. For example, simply asking a network to recognize the spatial layout of an image led to representations that transferred to popular vision tasks such as classification and detection \citep{doersch2015unsupervised,noroozi2016unsupervised}. Other works showed that prediction of color~\citep{zhang2016colorful,larsson2017colorization} and image orientation~\citep{gidaris2018unsupervised}, and invariance to data augmentation~\citep{dosovitskiy2014discriminative} can provide useful self-supervised tasks. Beyond single images, works have leveraged video cues such as object tracking~\citep{wang2015unsupervised}, frame ordering~\citep{misra2016shuffle}, and object boundary cues~\citep{li2016unsupervised,pathak2016learning}. Non-visual information can be equally powerful: information about camera motion~\citep{agrawal2015learning,jayaraman2015learning}, scene geometry~\citep{zamir2016generic}, or sound~\citep{arandjelovic2017look,arandjelovic2018objects} can all serve as natural sources of supervision.

While many of these tasks require predicting fixed quantities computed from the data, another class of \textit{contrastive} methods \citep{chopra2005learning, hadsell2006dimensionality} formulate their objectives in the learned representations themselves. CPC is a contrastive representation learning method that maximizes the mutual information between spatially removed latent representations with InfoNCE \citep{oord2018representation}, a loss function based on Noise-Contrastive Estimation \citep{gutmann2010noise, mnih2013learning}. Two other methods have recently been proposed using the same loss function, but with different associated prediction tasks. Contrastive Multiview Coding \citep{tian2019contrastive} maximizes the mutual information between representations of different views of the same observation. Augmented Multiscale Deep InfoMax (AMDIM, \citet{bachman2019learning}) is most similar to CPC in that it makes predictions across space, but differs in that it also predicts representations across layers in the model. Instance  Discrimination is another contrastive objective which encourages representations that can discriminate between individual examples in the dataset \citep{wu2018unsupervised}. 

A common alternative approach for improving data efficiency is \textbf{label-propagation} \citep{zhu2002labelprop}, where a classifier is trained on a subset of labeled data, then used to label parts of the unlabeled dataset. %, after which the process is repeated
This label-propagation can either be discrete (as in pseudo-labeling, \citet{lee2013pseudo}) or continuous (as in entropy minimization, \citet{grandvalet2005semi}). The predictions of this classifier are often constrained to be smooth with respect to certain deformations, such as data-augmentation \citep{xie2019unsupervised} or adversarial perturbation \citep{miyato2018virtual}. Representation learning and label propagation have been shown to be complementary and can be combined to great effect \citep{zhai2019s}, hence we focus solely on representation learning in this work.

\section{Results}
\label{sec:experiments}

When testing whether CPC enables data-efficient learning, we wish to use the best representative of this model class. Unfortunately, purely unsupervised metrics tell us little about downstream performance, and implementation details have been shown to matter enormously \citep{doersch2017multi, kolesnikov2019revisiting}. Since most representation learning methods have previously been evaluated using linear classification, we use this benchmark to guide a series of modifications to the training protocol and architecture (section \ref{sec:experiments-linear}) and compare to published results. In section \ref{sec:experiments-efficient} we turn to our central question of whether CPC enables data-efficient classification. Finally, in section \ref{sec:experiments-pascal} we investigate the generality of our results through transfer learning to PASCAL VOC 2007.

\subsection{From CPC v1 to CPC v2}
\label{sec:experiments-linear}
The overarching principle behind our new model design is to increase the scale and efficiency of the encoder architecture while also maximizing the supervisory signal we obtain from each image.  At the same time, it is important to control the types of predictions that can be made across image patches, by removing low-level cues which might lead to degenerate solutions. To this end, we augment individual patches independently using stochastic data-processing techniques from supervised and self-supervised learning. 

\begin{figure}
\centering
    \includegraphics[width=0.48\textwidth]{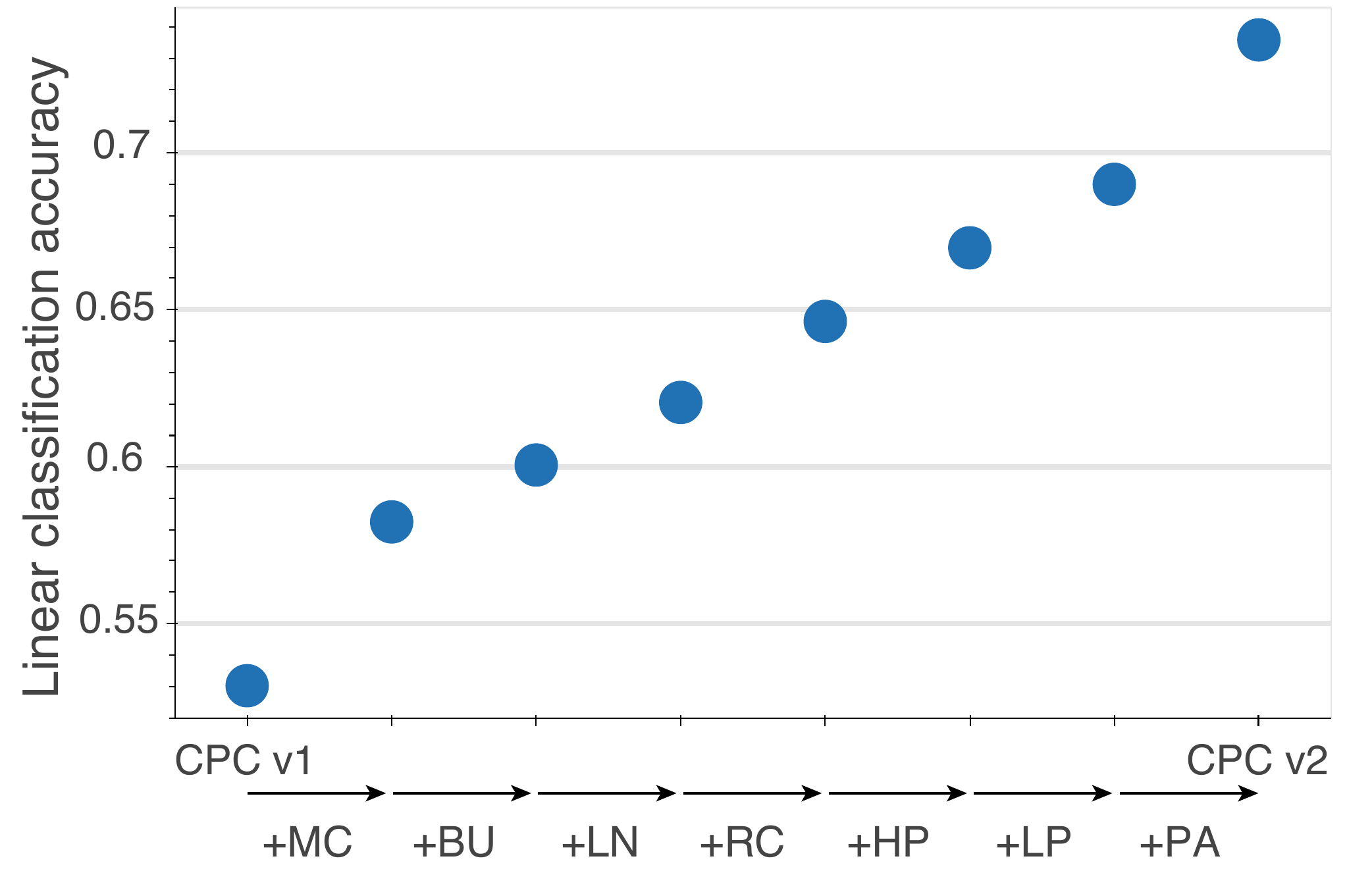}
    \vspace{-2em}
   \caption{Linear classification performance of new variants of CPC, which incrementally add a series of modifications. MC: model capacity. BU: bottom-up spatial predictions. LN: layer normalization. RC: random color-dropping. HP: horizontal spatial predictions. LP: larger patches. PA: further patch-based augmentation. Note that these accuracies are evaluated on a custom validation set and are therefore not directly comparable to the results we report on the official validation set.}
    \vspace{-1em}
\label{fig:linear_class}
\end{figure}

We identify four axes for model capacity and task setup that could impact the model's performance. The first axis increases model capacity by increasing depth and width, while the second improves training efficiency by introducing layer normalization. The third axis increases task complexity by making predictions in all four directions, and the fourth does so by performing more extensive patch-based augmentation.

\paragraph{Model capacity.} Recent work has shown that larger networks and more effective training improves self-supervised learning \citep{doersch2017multi, kolesnikov2019revisiting}, but the original CPC model used only the first 3 stacks of a ResNet-101 architecture.  Therefore, we convert the third residual stack of the  ResNet-101 (containing 23 blocks, 1024-dimensional feature maps, and 256-dimensional bottleneck layers) to use 46 blocks with 4096-dimensional feature maps and 512-dimensional bottleneck layers.  We call the resulting network ResNet-161. Consistent with prior results, this new architecture delivers better performance without any further modifications (Fig.\ \ref{fig:linear_class}, \textbf{+5\%} Top-1 accuracy). We also increase the model's expressivity by increasing the size of its receptive field with larger patches (from 64$\times$64 to 80$\times$80 pixels; \textbf{+2\%} Top-1 accuracy). 

\paragraph{Layer normalization.} Large architectures are more difficult to train efficiently. Early works on context prediction with patches used batch normalization~\citep{ioffe2015batch,doersch2015unsupervised} to speed up training. However, with CPC we find that batch normalization actually harms downstream performance of large models.  We hypothesize that batch normalization allows these models to find a trivial solution to CPC: it introduces a dependency between patches (through the batch statistics) that can be exploited to bypass the constraints on the receptive field. Nevertheless we find that we can reclaim much of batch normalization's training efficiency by using layer normalization (\textbf{+2\%} accuracy, \citet{ba2016layernorm}).

\begin{table}[t]
\caption{Linear classification accuracy, and comparison to other self-supervised methods. In all cases the feature extractor is optimized in an unsupervised manner, using one of the methods listed below. A linear classifier is then trained on top using all labels in the ImageNet dataset, and evaluated using a single crop. Prior art reported from [1] \citet{wu2018unsupervised}, 
[2] \citet{zhuang2019local}, 
[3] \citet{he2019momentum},
[4] \citet{misra2019pirl},
[5] \citet{doersch2017multi}, 
[6] \citet{kolesnikov2019revisiting}, 
[7] \citet{oord2018representation}, 
[8] \citet{donahue2019large}, 
[9] \citet{bachman2019learning}, 
[10] \citet{tian2019contrastive}. }
\label{tab:linear_class}
\begin{center}
\begin{small}
\begin{sc}
\begin{tabular}{lccc}
\toprule
Method & Params (M) & Top-1 & Top-5 \\
\midrule \\ 
\multicolumn{4}{l}{\textit{Methods using ResNet-50:}} \\
Instance Discr. [1] & 24 & 54.0 & - \\
Local Aggr. [2]  & 24 & 58.8 & - \\
MoCo [3] & 24 & 60.6 & - \\ 
PIRL [4] & 24 & 63.6 & - \\
\midrule
CPC v2 - ResNet-50 & 24 & \textbf{63.8} & \textbf{85.3} \\ 
\midrule \\ 
\multicolumn{4}{l}{\textit{Methods using different architectures:}} \\
Multi-task [5]  & 28 & - & 69.3 \\
Rotation [6] & 86 & 55.4 & - \\ 
CPC v1  [7] & 28 & 48.7 & 73.6 \\ 
BigBiGAN [8] & 86 & 61.3 & 81.9 \\
AMDIM  [9] & 626 & 68.1 & - \\ 
CMC [10]  & 188 & 68.4 & 88.2 \\ 
MoCo [2]  & 375 & 68.6 & - \\ 
\midrule
CPC v2 - ResNet-161 & 305 & \textbf{71.5} & \textbf{90.1} \\ 
\bottomrule
\end{tabular}
\end{sc}
\end{small}
\end{center}
\vskip -0.1in
\end{table}

\paragraph{Prediction lengths and directions.} Larger architectures also run a greater risk of overfitting. We address this by asking more from the network: specifically, whereas the model in  \citet{oord2018representation} predicted each patch using only context from above, we repeatedly predict the same patch using context from below, the right and the left (using separate context networks), resulting in up to four times as many prediction tasks. Additional predictions tasks incrementally increased accuracy (adding bottom-up predictions: \textbf{+2\%} accuracy; using all four spatial directions: \textbf{+2.5\%} accuracy).

\begin{table*}[t]
\caption{Data-efficient image classification. We compare the accuracy of two ResNet classifiers, one trained on the raw image pixels, the other on the proposed CPC v2 features, for varying amounts of labeled data. Note that we also fine-tune the CPC features for the supervised task, given the limited amount of labeled data. Regardless, the ResNet trained on CPC features systematically surpasses the one trained on pixels, even when given 2--5$\times$ less labels to learn from. The red (respectively, blue) boxes highlight comparisons between the two classifiers, trained with different amounts of data, which illustrate a 5$\times$ (resp. 2$\times$) gain in data-efficiency in the low-data (resp. high-data) regime.}
\label{tab:efficient1}
\vskip 0.15in
\begin{center}
\begin{small}
\begin{sc}
\begin{tabular}{l c c c c c c c }
\toprule
    Labeled data & 1\% & 2\% & 5\% & 10\% & 20\%  & 50\% & 100\% \\
    \midrule 
     &  \multicolumn{7}{c}{Top-1 accuracy} \\ 
    ResNet-200 trained on pixels & 23.1 & 34.8 & \colorboxed{red}{50.6} & 62.5 & 70.3 & 75.9 & \colorboxed{blue}{80.2} \\
    ResNet-33 trained on CPC features & \colorboxed{red}{52.7} & 60.4 & 68.1 & 73.1 &76.7 & \colorboxed{blue}{81.2} & 83.4 \\
    Gain in data-efficiency &  {\bf \color{red}5$\mathbf{\times}$} & 2.5$\times$ & 2$\times$ & 2$\times$ & 2.5$\times$ & {\bf \color{blue}2$\mathbf{\times}$} &   \\
    \midrule 
     &  \multicolumn{7}{c}{Top-5 accuracy} \\ 
    ResNet-200 trained on pixels & 44.1 & 59.9 & \colorboxed{red}{75.2} & 83.9 & 89.4 & 93.1 & \colorboxed{blue}{95.2} \\
    ResNet-33 trained on CPC features & \colorboxed{red}{78.3} & 83.9 & 88.8 & 91.2 & 93.3 & \colorboxed{blue}{95.6} & 96.5 \\
    Gain in data-efficiency & {\bf \color{red}5$\mathbf{\times}$} & 5$\times$ & 2$\times$ & 2.5$\times$ & 2$\times$ & {\bf \color{blue}2$\mathbf{\times}$} &  \\
\bottomrule
\end{tabular}
\end{sc}
\end{small}
\end{center}
\end{table*}

\paragraph{Patch-based augmentation.} 
If the network can solve CPC using low-level patterns (e.g.\ straight lines continuing between patches or chromatic aberration), it need not learn semantically meaningful content. Augmenting the low-level variability across patches can remove such cues. To that effect, the original CPC model spatially jittered individual patches independently.  We further this logic by adopting the `color dropping' method of~\citet{doersch2015unsupervised}, which randomly drops two of the three color channels in each patch, and find it to deliver systematic gains (\textbf{+3\%} accuracy). We therefore continued by adding a fixed, generic augmentation scheme using the primitives from \citet{cubuk2018autoaugment} (e.g.\ shearing, rotation, etc), as well as random elastic deformations and color transforms (\citet{de2018clinically}, \textbf{+4.5\%} accuracy in total). Note that these augmentations introduce some inductive bias about content-preserving transformations in images, but we do not optimize them for downstream performance (as in \citet{cubuk2018autoaugment} and \citet{lim2019fast}).

\paragraph{Comparison to previous art.} Cumulatively, these fairly straightforward implementation changes lead to a substantial improvement to the original CPC model, setting a new state-of-the-art in linear classification of 71.5\% Top-1 accuracy (compared to 48.7\% for the original, see table \ref{tab:linear_class}). Note that our architecture differs from ones used by other works in self-supervised learning, while using a number of parameters which is comparable to recently-used ones. The great diversity of network architectures (e.g.\ BigBiGAN employs a RevNet-50 with a  $\times4$ widening factor, AMDIM a customized ResNet architecture, CMC a ResNet-50 $\times2$ and Momentum Contrast and ResNet-50 $\times$4) make any apples-to-apples comparison with these works challenging. In order to compare with published results which use the same architecture, we therefore also trained a ResNet-50 architecture for the CPC v2 objective, arriving at 63.8\% linear classification accuracy. This model outperforms methods which use the same architecture, as well as many recent approaches which at times use substantially larger ones \citep{doersch2017multi, oord2018representation, kolesnikov2019revisiting, zhuang2019local, donahue2019large}. 

\subsection{Efficient image classification}
\label{sec:experiments-efficient}
We now turn to our original question of whether CPC can enable data-efficient image recognition. 

\paragraph{Supervised baseline.} We start by evaluating the performance of purely-supervised networks as the size of the labeled dataset $\sD_l$ varies from 1\% to 100\% of ImageNet, training separate classifiers on each subset. We compared a range of different architectures (ResNet-50, -101, -152, and -200) and found a ResNet-200 to work best across all data-regimes. After tuning the supervised model for low-data classification (varying network depth, regularization, and optimization parameters) and extensive use of data-augmentation (including the transformations used for CPC pre-training), the accuracy of the best model reaches 44.1\% Top-5 accuracy when trained on 1\% of the dataset (compared to 95.2\% when trained on the entire dataset, see Table \ref{tab:efficient1} and Fig.\ \ref{fig:intro}, red).

\paragraph{Contrastive Predictive Coding.} We now address our central question of whether CPC enables data-efficient learning. We follow the same paradigm as for the supervised baseline (training and evaluating a separate classifier for each labeled subset), stacking a neural network classifier on top of the CPC latents $\vz=f_\theta(\vx)$ rather than the raw image pixels $\vx$. Specifically, we stack an 11-block ResNet classifier $h_\psi$ on top of the 14$\times$14 grid of CPC latents, and train it using the same protocol as the supervised baseline (see section \ref{sec:evaluation-protocol}). During an initial phase we keep the CPC feature extractor fixed and train the ResNet classifier till convergence (see Table \ref{tab:efficient2} for its performance). We then fine-tune the entire stack $h_\psi \circ f_\theta$ for the supervised objective, for a small number of epochs (chosen by cross-validation). In Table \ref{tab:efficient1} and Fig.\ \ref{fig:intro} (blue curve) we report the results of this fine-tuned model. 

This procedure leads to a substantial increase in accuracy, yielding 78.3\% Top-5 accuracy with only 1\% of the labels, a 34\% absolute improvement (77\% relative) over purely-supervised methods. Surprisingly, when given the entire dataset, this classifier reaches 83.4\%/96.5\% Top1/Top5 accuracy, surpassing our supervised baseline (ResNet-200: 80.2\%/95.2\% accuracy) and published results (original ResNet-200 v2: 79.9\%/95.2\%, \citet{he2016identity}; with AutoAugment:  80.0\%/95.0\%,  \citet{cubuk2018autoaugment}). Using this representation also leads to gains in data-efficiency. With only 50\% of the labels our classifier surpasses the supervised baseline given the entire dataset, representing a 2$\times$ gain in data-efficiency (see table \ref{tab:efficient1}, blue boxes). Similarly, with only 1\% of the labels, our classifier surpasses the supervised baseline given 5\% of the labels (i.e.\ a 5$\times$ gain in data-efficiency, see table \ref{tab:efficient1}, red boxes).

Note that we are comparing two different model \textit{classes} as opposed to specific models or instantiations of these classes. As result we have searched for the best representative of each class, landing on the ResNet-200 for purely supervised ResNets and our wider ResNet-161 for CPC pre-training (with a ResNet-33 for downstream classification). Given the difference in capacity between these models (the ResNet-200 has approximately 60 million parameters whereas our combined model has over 500 million parameters), we verified that supervised learning would not benefit from this larger architecture. Training the ResNet-161 + ResNet-33 stack (including batch normalization throughout) in a purely supervised manner yielded results that were similar to that of the ResNet-200 (80.3\%/95.2\% Top-1/Top-5 accuracy). This result is to be expected: the family of ResNet-50, -101, and -200 architectures are designed for supervised learning, and their capacity is calibrated for the amount of training signal present in ImageNet labels; larger architectures only run a greater risk of overfitting. In contrast, the CPC training objective is much richer and requires larger architectures to be taken advantage of, as evidenced by the difference in linear classification accuracy between a ResNet-50 and ResNet-161 trained for CPC (table 1, 63.8\% vs 71.5\% Top-1 accuracy). 

\begin{table}[t]
\caption{Comparison to other methods for semi-supervised learning. \textit{Representation learning} methods use a classifier to discriminate an unsupervised representation, and optimize it solely with respect to labeled data. \textit{Label-propagation} methods on the other hand further constrain the classifier with smoothness and entropy criteria on unlabeled data, making the additional assumption that all training images fit into a single (unknown) testing category. When evaluating CPC v2, BigBiGAN, and AMDIM, we train a ResNet-33 on top of the representation, while keeping the representation \textit{fixed} or allowing it to be \textit{fine-tuned}. All other results are reported from their respective papers: [1] \citet{zhai2019s}, [2] \citet{xie2019unsupervised}, [3] \citet{wu2018unsupervised}, [4] \citet{misra2019pirl}.}
\label{tab:efficient2}
\vskip 0.15in
\begin{center}
\begin{small}
\begin{sc}
\begin{tabular}{l c c c }
\toprule
    Labeled data & 1\% & 10\% &  100\% \\
    \midrule
      & \multicolumn{3}{c}{Top-5 accuracy} \\ 
    % \midrule \\ 
    Supervised baseline & 44.1 & 83.9 & 95.2 \\ 
    \midrule \\
    \multicolumn{3}{l}{\textit{Methods using label-propagation:}} \\
    Pseudolabeling [1] & 51.6 &  82.4  & - \\    
    VAT + Entropy Min. [1] & 47.0  & 83.4  & -\\ 
    Unsup. Data Aug. [2]  & -  & 88.5  & -\\ 
    Rot. + VAT + Ent. Min. [1]  & -  &\textbf{91.2} & 95.0 \\ 
    \midrule \\
    \multicolumn{3}{l}{\textit{Methods using representation learning only:}} \\
    Instance Discr. [3]& 39.2 & 77.4  & -\\
    PIRL [4] & 57.2 & 83.8 & - \\ 
    Rotation [1]  & 57.5 & 86.4  & - \\ 
  BigBiGAN (fixed) & 55.2 & 78.8  & 87.0 \\ 
  AMDIM (fixed)  & 67.4  & 85.8  & 92.2 \\
    \midrule 
    CPC v2 (fixed)  &  77.1  & 90.5  & 96.2 \\ 
    CPC v2 (fine-tuned)  & \textbf{78.3}  & \textbf{91.2} & \textbf{96.5}\\
\bottomrule
\end{tabular}
\end{sc}
\end{small}
\end{center}
\end{table}

\paragraph{Other unsupervised representations.} How well does the CPC representation compare to other representations that have been learned in an unsupervised manner? Table~\ref{tab:efficient2} compares our best model with other works on efficient recognition. We consider three objectives from different model classes: self-supervised learning with rotation prediction \citep{zhai2019s}, large-scale adversarial feature learning (BigBiGAN, \citet{donahue2019large}), and another contrastive prediction objective (AMDIM, \citet{bachman2019learning}). \citet{zhai2019s} evaluate the low-data classification performance of representations learned with rotation prediction using a similar paradigm and architecture (ResNet-152 with a $\times$2 widening factor), hence we report their results directly: given 1\% of ImageNet labels, their method achieves 57.5\% Top-5 accuracy. The authors of BigBiGAN and AMDIM do not report results on efficient classification, hence we evaluated these representations using the same paradigm we used for evaluating CPC. Specifically, since fine-tuned representations yield only marginal gains over fixed ones (e.g.\ 77.1\% vs 78.3\% Top-5 accuracy given 1\% of the labels, see table \ref{tab:efficient2}), we train an identical ResNet classifier on top of these representations while keeping them fixed. Given 1\% of ImageNet labels, classifiers trained on top of BigBiGAN and AMDIM achieve 55.2\% and 67.4\% Top-5 accuracy, respectively.

Finally, Table~\ref{tab:efficient2} (top) also includes results for label-propagation algorithms. Note that the comparison is imperfect: these methods have an advantage in assuming that all unlabeled images can be assigned to a single category.  At the same time, prior works (except for \citet{zhai2019s} which use a ResNet-50 $\times$4) report results with smaller networks, which may degrade performance relative to ours. Overall, we find that our results are on par with or surpass even the strongest such results~\citep{zhai2019s}, even though this work combines a variety of techniques (entropy minimization, virtual adversarial training, self-supervised learning, and pseudo-labeling) with a large architecture whose capacity is similar to ours.

In summary, we find that CPC provides gains in data-efficiency that were previously unseen from representation learning methods, and rival the performance of the more elaborate label-propagation algorithms. 

\subsection{Transfer learning: image detection on PASCAL VOC 2007}
\label{sec:experiments-pascal}
We next investigate transfer learning performance on object detection on the PASCAL VOC 2007 dataset, which reflects the practical scenario where a representation must be trained on a dataset with different statistics than the dataset of interest. This dataset also tests the efficiency of the representation as it only contains 5011 labeled images to train from. The standard protocol in this setting is to train an ImageNet classifier in a supervised manner, and use it as a feature extractor for a Faster-RCNN object detection architecture \citep{ren2015faster}. Following this procedure, we obtain 74.7\% mAP with a ResNet-152 (Table \ref{tab:pascal_det}). In contrast, if we use our CPC encoder as a feature extractor in the same setup, we obtain 76.6\% mAP. This represents one of the first results where unsupervised pre-training surpasses supervised pre-training for transfer learning. Note that consistently with the previous section, we limit ourselves to comparing the two model \textit{classes} (supervised vs. self-supervised), choosing the best architecture for each. Concurrently with our results, \citet{he2019momentum} achieve 74.9\% in the same setting.

\begin{table}[t]
\caption{Comparison of PASCAL VOC 2007 object detection accuracy to other transfer methods. The supervised baseline learns from the entire labeled ImageNet dataset and fine-tunes for PASCAL detection. The second class of methods learns from the same \emph{unlabeled} images before transferring. The architecture column specifies the object detector (Fast-RCNN or Faster-RCNN) and the feature extractor (ResNet-50, -101, -152, or -161). All of these methods pre-train on the ImageNet dataset, except for DeeperCluster which learns from the larger, but uncurated, YFCC100M dataset \citep{thomee2015yfcc100m}. All methods fine-tune on the PASCAL 2007 training set, and are evaluted in terms of mean average precision (mAP). Prior art reported from [1] \citet{dosovitskiy2014discriminative}, [2] \citet{doersch2017multi}, [3] \citet{pathak2016learning},  [4] \citet{zhang2016colorful},  [5] \citet{doersch2015unsupervised},  [6] \citet{wu2018unsupervised}, [7] \citet{caron2018deep}, [8] \citet{caron2019leveraging}, [9] \citet{zhuang2019local}, [10] \citet{misra2019pirl} [11] \citet{he2019momentum}.} 
\label{tab:pascal_det}
\begin{center}
\begin{small}
\begin{sc}
\begin{tabular}{l l c}
\toprule
Method  & Architecture & mAP \\ 
\midrule \\
    \textit{Transfer using labeled data:} & & \\
    Supervised baseline & Faster: R152 & 74.7 \\
    \midrule \\
    \textit{Transfer using unlabeled data:} & \\
    Exemplar [1] \small{by} [2] & Faster: R101 & 60.9  \\
    Motion Segm. [3] \small{by} [2] & Faster: R101 & 61.1  \\
    Colorization [4] \small{by} [2] & Faster: R101 & 65.5 \\
    Relative Pos. [5] \small{by} [2]   & Faster: R101 & 66.8 \\ 
    Multi-task [2] & Faster: R101 & 70.5  \\
    Instance Discr. [6] & Faster: R50 & 65.4 \\ 
    Deep Cluster [7] & Fast: VGG-16 & 65.9 \\
    Deeper Cluster [8] & Fast: VGG-16 & 67.8  \\
    Local Aggregation [9] & Faster: R50 & 69.1 \\ 
    PIRL [10] & Faster: R50 &  73.4 \\ 
    Momentum Contrast [11] & Faster: R50 & 74.9 \\ 
    \midrule
    CPC v2 & Faster: R161 & \textbf{76.6} \\
    \bottomrule
\end{tabular}
\end{sc}
\end{small}
\end{center}
\end{table}

\section{Discussion}

We asked whether CPC could enable data-efficient image recognition, and found that it indeed greatly improves the accuracy of classifiers and object detectors when given small amounts of labeled data.  Surprisingly, CPC even improves their peformance when given ImageNet-scale labels. Our results show that there is still room for improvement using relatively straightforward changes such as augmentation, optimization, and network architecture. Overall, these results open the door toward research on problems where data is naturally limited, e.g.\ medical imaging or robotics.

Furthermore, images are far from the only domain where unsupervised representation learning is important: for example, unsupervised learning is already a critical step in natural language processing~\citep{mikolov2013word2vec,devlin2018bert}, and shows promise in domains like audio~\citep{oord2018representation,arandjelovic2018objects,arandjelovic2017look}, video~\citep{jing2018self,misra2016shuffle}, and robotic manipulation~\citep{pinto2016supersizing,pinto2016supervision,sermanet2018time}.  
Currently much self-supervised work builds upon tasks tailored for a specific domain (often images), which may not be easily adapted to other domains.  Contrastive prediction methods, 
including the techniques proposed in this paper, are task agnostic and could therefore serve as a unifying framework for integrating these tasks and modalities. This generality is particularly useful given that many real-world environments are inherently multimodal, e.g.\ robotic environments which can have vision, audio, touch, proprioception, action, and more over long temporal sequences. Given the importance of increasing the amounts of self-supervision (via additional prediction tasks), integrating these modalities and tasks could lead to unsupervised representations which rival the efficiency and effectiveness of human ones. 

\bibliography{bibliography}
\bibliographystyle{icml2020}

\clearpage
\appendix
\twocolumn[\icmltitle{Supplementary Information: \\ Data-Efficient Image Recognition with Contrastive Predictive Coding}]
\section{Self-supervised pre-training}
\label{sec:appendix_pretrain}

\paragraph{Model architecture:} Having extracted 80$\times$80 patches with a stride of 36$\times$36 from an input image with 260$\times$260 resolution, we end up with a grid of 6$\times$6 image patches. We transform each one with a ResNet-161 encoder which terminates with a mean pooling operation, resulting in a [6,6,4096] tensor representation for each image. We then aggregate these \lstinline[language=Python]{latents} into a 6$\times$6 grid of context vectors, using a \lstinline[language=Python]{pixelCNN}. We use this context to make the predictions and compute the \lstinline[language=Python]{CPC} loss. 

\begin{lstlisting}[language=Python, numbers=none, framexleftmargin=0mm]
def pixelCNN(latents):
    # latents: [B, H, W, D]
    cres = latents
    cres_dim = cres.shape[-1]
    for _ in range(5):
      c = Conv2D(output_channels=256,
                 kernel_shape=(1, 1))(cres)
      c = ReLU(c)
      c = Conv2D(output_channels=256,
                 kernel_shape=(1, 3))(c)
      c = Pad(c, [[0, 0], [1, 0], [0, 0], [0, 0]])
      c = Conv2D(output_channels=256,
                 kernel_shape=(2, 1),
                 type='VALID')(c)
      c = ReLU(c)
      c = Conv2D(output_channels=cres_dim,
                 kernel_shape=(1, 1))(c)
      cres = cres + c
    cres = ReLU(cres)
    return cres

def CPC(latents, target_dim=64, emb_scale=0.1,
        steps_to_ignore=2, steps_to_predict=3):
    # latents: [B, H, W, D]
    loss = 0.0
    context = pixelCNN(latents)
    targets = Conv2D(output_channels=target_dim, 
                     kernel_shape=(1, 1))(latents)
    batch_dim, col_dim, rows = targets.shape[:-1]
    targets = reshape(targets, [-1, target_dim])
    for i in range(steps_to_ignore, steps_to_predict):
      col_dim_i = col_dim - i - 1
      total_elements = batch_dim * col_dim_i * rows
    
      preds_i = Conv2D(output_channels=target_dim,
                      kernel_shape=(1, 1))(context)
      preds_i = preds_i[:, :-(i+1), :, :] * emb_scale
      preds_i = reshape(preds_i, [-1, target_dim])
      
      logits = matmul(preds_i, targets, transp_b=True)
    
      b = range(total_elements) / (col_dim_i * rows)
      col = range(total_elements) % (col_dim_i * rows)
      labels = b * col_dim * rows + (i+1) * rows + col
    
      loss += cross_entropy_with_logits(logits, labels)
    return loss
\end{lstlisting}

\paragraph{Image preprocessing: } The final CPC v2 image processing pipeline we adopt consists of the following steps. We first resize the image to 300$\times$300 pixels and randomly extract a 260$\times$260 pixel crop, then divide this image into a 6$\times$6 grid of 80$\times$80 patches. Then, for every patch: 
\begin{enumerate}
    \item Randomly choose two transformations from \citet{cubuk2018autoaugment} and apply them using default parameters.
    \item Using the primitives from \citet{de2018clinically}, randomly apply elastic deformation and shearing with a probability of 0.2. Randomly apply their color-histogram automentations with a probability of 0.2.
    \item Randomly apply the color augmentations from \citet{szegedy2014going} with a probability of 0.8. 
    \item Randomly project the image to grey-scale with a probability of 0.25. 
\end{enumerate}

\paragraph{Optimization details:} We train the network for the CPC objective using the Adam  optimizer \citep{kingma2014adam} for 200 epochs, using a learning rate of 0.0004, $\beta_1 = 0.8$, $\beta_2 = 0.999$, $\epsilon=10^{-8}$ and Polyak averaging with a decay of 0.9999. We also clip gradients to have a maximum norm of 0.01. We train the model with a batch size of 512, which we spread across 32 workers. 

\section{Linear classification}
\label{sec:appendix_linear}

\paragraph{Model architecture:} For linear classification we encode each image in the same way as during self-supervised pre-training (section \ref{sec:appendix_pretrain}), yielding a 6$\times$6 grid of 4096-dimensional features vectors. We then use Batch-Normalization \citep{ioffe2015batch} to normalize the features (omitting the scale parameter) followed by a 1$\times$1 convolution mapping each feature in the grid to the 1000 logits for ImageNet classification. We then spatially mean-pool these logits to end up with the final log probabilities for the linear classification.

\paragraph{Image preprocessing: } We use the same data pipeline as for self-supervised pre-training (section \ref{sec:appendix_pretrain}). 

\paragraph{Optimization details:} We use the Adam optimizer with a learning rate of 0.0005. We train the model with a batch size of 512 images spread over 16 workers.

\section{Efficient classification}
\subsection{Purely supervised}
\label{sec:appendix_efficient_sup}

\paragraph{Model architecture:} We investigate using ResNet-50, ResNet-101, ResNet-152, and ResNet-200 model architectures, all of them using the `v2' variant \citep{he2016identity}, and find larger architectures to perform better, even when given smaller amounts of data. We insert a DropOut layer before the final linear classification layer \citep{srivastava2014dropout}. 

\paragraph{Image preprocessing: } We extract a randomly sized crop, as in the augmentations of \citet{szegedy2014going}. We follow this with the same image transformations as for self-supervised pre-training (steps 1--4).
    
\paragraph{Optimization details:} We use stochastic gradient descent, varying the learning rate in \{0.05, 0.1, 0.2\}, the weight decay  logarithmically from $10^{-5}$ to $10^{-2}$, the DropOut linearly from 0 to 1, and the batch size per worker in \{16, 32\}. We search for the best-performing model separately for each subset of labeled training data, as more labeled data requires less regularization. Having chosen these hyperparameters using a separate validation set (approximately 10k images which we remove from the training set), we evaluate each model on the test set (i.e. the publicly available ILSVRC-2012 validation set). 

\subsection{Semi-supervised with CPC}

\paragraph{Model architecture:} We apply the CPC encoder directly to the image, resulting in a 14$\times$14 grid of feature vectors. These features are used as inputs to an 11-block ResNet classifier with 4096-dimensional hiddens layers and 1024-dimensional bottleneck layers. As for the supervised baseline, we insert DropOut after the final mean-pooling operation and before the final linear classifier. 

\paragraph{Image preprocessing:} We use the same pipeline as the supervised baseline. 

\paragraph{Optimization details:} We start by training the classifier while keeping the CPC features fixed. To do so we search through the same set of hyperparameters as the supervised baseline. After training the classifier till convergence, we fine-tune the entire stack for classification. In this phase we keep the optimization details of each component the same as previously: the classifier is fine-tuned with SGD, while the encoder is fine-tuned with Adam.
% \bibliography{bibliography}
% \bibliographystyle{icml2020}


\end{document}